\documentclass[letterpaper, 10 pt, conference]{ieeeconf}  

\IEEEoverridecommandlockouts                              

\usepackage{amsmath} 
\usepackage{amssymb} 

\usepackage{amsthm}

\usepackage{graphicx}
\usepackage{makecell}
\usepackage{threeparttable} 
\usepackage{subcaption}
\usepackage{xcolor}
\usepackage{booktabs}
\usepackage{multirow}

\usepackage[pagebackref = false,breaklinks,colorlinks,citecolor=green]{hyperref}

\usepackage{xurl}

\let\titleold\title
\renewcommand{\title}[1]{\titleold{#1}\newcommand{\thetitle}{#1}}

\AtEndPreamble{
    \usepackage[capitalize]{cleveref}
    \crefname{section}{Sec.}{Secs.}
    \Crefname{section}{Section}{Sections}
    \Crefname{table}{Table}{Tables}
    \crefname{table}{Tab.}{Tabs.}
}

\newif\ifproofread
\proofreadfalse

\title{\LARGE \bf
Unleashing the Power of Discrete-Time State Representation: \\Ultrafast Target-based IMU-Camera Spatial-Temporal Calibration
}

\author{Junlin Song, Antoine Richard, and Miguel Olivares-Mendez
\thanks{Space Robotics (SpaceR) Research Group, Int. Centre for Security, Reliability and Trust (SnT), University of Luxembourg, Luxembourg.} 
}

\begin{document}

\maketitle
\thispagestyle{empty}
\pagestyle{empty}

\begin{abstract}

Visual-inertial fusion is crucial for a large amount of intelligent and autonomous applications, such as robot navigation and augmented reality. To bootstrap and achieve optimal state estimation, the spatial-temporal displacements between IMU and cameras must be calibrated in advance. Most existing calibration methods adopt continuous-time state representation, more specifically the B-spline. Despite these methods achieve precise spatial-temporal calibration, they suffer from high computational cost caused by continuous-time state representation. To this end, we propose a novel and extremely efficient calibration method that unleashes the power of discrete-time state representation. Moreover, the weakness of discrete-time state representation in temporal calibration is tackled in this paper. With the increasing production of drones, cellphones and other visual-inertial platforms, if one million devices need calibration around the world, saving one minute for the calibration of each device means saving 2083 work days in total. To benefit both the research and industry communities, the open-source implementation is released at \url{https://github.com/JunlinSong/DT-VI-Calib}.

\end{abstract}



\section{Introduction}

State estimation is a fundamental research topic in the robotics and computer vision communities. There have been tremendous advances over the past few decades, from single sensor to multi-sensor fusion. In practice, the use of a single sensor may be limited by inherent flaws, such as scale ambiguity in monocular simultaneous localization and mapping (SLAM). Different sensors can complement each other, thus significantly improving the overall localization and perception capability. Among different multi-sensor fusion schemes, visual-inertial fusion has attracted great attention, as visual-inertial sensor suite has several advantages: small size, low power consumption, and low cost. Nowadays, visual-inertial odometry (VIO) is widely used in AR/VR \cite{engel2023project, de2025monado}, robotics \cite{delmerico2018benchmark, hanover2024autonomous}, and planetary exploration \cite{delaune2021range, alberico2024structure}.

\begin{figure}[htbp]
  \centering
    \begin{subfigure}{0.23\textwidth}
        \centering
        \includegraphics[width=\textwidth, height=0.8\textwidth]{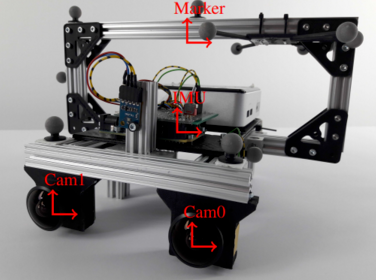}
        \caption{}
        \label{Frames in TUM-VI}
    \end{subfigure}
    \hfill
    \begin{subfigure}{0.23\textwidth}
        \centering
        \includegraphics[width=\textwidth, height=0.8\textwidth]{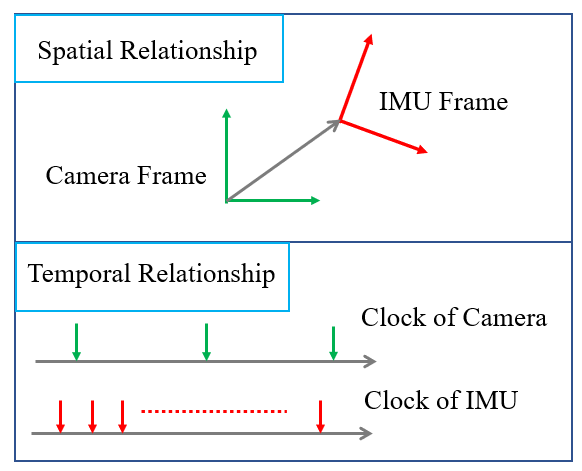}
        \caption{}
        \label{spatial-temporal}
    \end{subfigure}
  \caption{(a) Stereo visual-inertial sensor prototype of the TUM-VI dataset \cite{schubert2018tum}. (b) The spatial-temporal relationship between IMU and camera.}
  \label{fig: vi-calib}
\end{figure}

The successful running of a VIO system relies on accurate spatial-temporal calibration for IMU and cameras (see Fig. \ref{fig: vi-calib}). Spatial calibration plays the role of aligning the coordinate frames for different sensor measurements. Temporal calibration aligns different clocks that timestamp the measurements. Temporal calibration is especially critical when strict hardware synchronization is unavailable.

\begin{table}
  \caption{State dimensions comparison of different calibration methods on the EuRoC \cite{burri2016euroc} and TUM-VI \cite{schubert2018tum} calibration sequences. For TUM-VI dataset, $imu1$ is used as an example sequence here. Image frequency is decreased from 20hz to 10hz, and 5hz.}
  \centering
  \aboverulesep=0ex
  \belowrulesep=0ex
  \scalebox{1.05}{
  \begin{tabular}{@{}c|c|c|ccc@{}}
    \toprule
    \multirow{2}{*}{Dataset} & \multirow{2}{*}{Duration (s)} & \multirow{2}{*}{Methods} & \multicolumn{3}{c}{Image frequency (hz)} \\ 
    \cmidrule(lr){4-6} & & & {20} & {10} & {5} \\
    \midrule
    \multirow{3}{*}{EuRoC} & \multirow{3}{*}{71.9} & Kalibr \cite{furgale2013unified} & 64888 & 64888  & 64888   \\
    & & Basalt \cite{sommer2020efficient} & 43256 & 43256  & 43256    \\
    & & Ours & 12747 & 6375  & 3198    \\
    \hline
    \multirow{3}{*}{TUM-VI} & \multirow{3}{*}{51.9} & Kalibr \cite{furgale2013unified} & 46846 & 46846  & 46846   \\
    & & Basalt \cite{sommer2020efficient} & 31232 & 31232  & 31232    \\
    & & Ours & 9345 & 4683  & 2352    \\
    \bottomrule
  \end{tabular}}
  \label{tab: state_dim}
\end{table}

\begin{table*}
  \caption{Comparisons between representative offline IMU-Camera calibration methods and our method.}
  \centering
  \begin{threeparttable}
  \begin{tabular}{@{}ccccccc@{}}
    \toprule
    Method & Year & Continuous? & Target-based? & Constant IMU biases? & Open-Source? & Optimization efficiency  \\
    \midrule
    Kalibr\cite{furgale2013unified} & 2013 & Continuous & Target-based & No & Yes & baseline \\
    \hline
    Basalt \cite{sommer2020efficient} & 2020 & Continuous & Target-based & Yes&  Yes & approximately 10x faster than Kalibr \\
    \hline
    MVIS \cite{yang2024multi} & 2024 & Discrete & Target-based & No & No \tnote{a} & approximately 3x faster than Kalibr \\
    \hline
    iKalibr \cite{chen2025ikalibr} & 2025 & Continuous & Target-less & Yes & Yes & Slower than Kalibr due to heavy SfM \cite{schonberger2016structure} \\
    \hline\hline
    Ours & 2026 & Discrete & Target-based & Yes & Yes & approximately 600x faster than Kalibr \\
    \bottomrule
  \end{tabular}
    \begin{tablenotes}
        \item[a] The relevant code of MVIS \cite{yang2024multi} (\url{https://github.com/yangyulin/mvis.git}) is unavailable.
    \end{tablenotes}
  \end{threeparttable}
  \label{tab: offline_calib}
\end{table*}

To address the calibration problem for IMU and cameras, extensive studies have been conducted in the literature, from theory to practice. Currently, almost all open-source IMU-Camera calibration methods employ a continuous-time state representation based on the B-spline. This type of method can obtain accurate and consistent calibration results with the aid of a calibration target, and the representative work is termed \textbf{Kalibr}, developed by \cite{furgale2013unified}. However, Kalibr suffers from high computational cost due to its B-spline based state representation. To reduce computational complexity, \cite{sommer2020efficient} further derive a novel and efficient derivative calculation method for the B-spline on Lie groups \cite{sola2018micro}. As IMU-Camera calibration method proposed by \cite{sommer2020efficient} has been integrated into Basalt project \cite{usenko2018double, usenko2019visual, sommer2020efficient}, hereafter, we use \textbf{Basalt} to refer to the calibration method presented in \cite{sommer2020efficient}.

Discrete-time state representation can also be applied to the target-based IMU-Camera spatial-temporal calibration task. Surprisingly, there are few works performing calibration in this way. Compared to continuous-time state representation, discrete-time state representation actually showcases even better state estimation performance on some well-known public benchmarks, for example, the odometry leaderboard\footnote{\scriptsize \url{https://www.cvlibs.net/datasets/kitti/eval_odometry.php}} of KITTI \cite{geiger2012we}, and the SLAM challenge leaderboard\footnote{\scriptsize \url{https://hilti-challenge.com/leader-board-2023.html}} of HILTI  \cite{10582435}. Those excellent works in leaderboards demonstrate that discrete-time state representation is still promising.

Discrete-time state representation is typically considered difficult or inferior for temporal calibration \cite{furgale2013unified, cioffi2022continuous, talbot2025continuous}. For example, the authors of Kalibr \cite{furgale2013unified} are concerned that discrete-time state representation requires a new state at each measurement time, which could be challenging for the utilization of high-frequency IMU measurements, and subsequent estimator design for temporal calibration.

This concern can be tackled with IMU preintegration \cite{forster2016manifold} for discrete-time formulation. Multiple IMU measurements between two consecutive images are aggregated as one pseudo-measurement, thus greatly reducing the state dimensions for optimization (see TABLE \ref{tab: state_dim}). However, naively utilizing existing IMU preintegration with Euler integration \cite{forster2016manifold, usenko2019visual, delama2024equivariant, yang2024multi} may lead to worse temporal calibration than continuous-time state representation. This is due to the unsatisfactory accuracy of IMU integration, as shown in Section \ref{sec: results}. Therefore, we design a higher-order integration scheme to address this limitation.

MVIS \cite{yang2024multi} is a recent work that adopts a discrete-time state representation for IMU-Camera calibration. However, due to the use of a gravity-aligned reference frame, MVIS sacrifices efficiency by introducing additional multiple 3D feature positions in the state vector. Instead, by enabling joint gravity estimation with the IMU preintegration model, our method eliminates all these features from the state vector. Furthermore, redundant IMU biases are removed. These differences allow our method to fully unleash the efficiency power of discrete-time calibration (see TABLE \ref{tab: offline_calib}).

Key contributions of this work can be summarized as:

\begin{itemize}

\item Our novel IMU-Camera calibration method jointly determines the spatial-temporal calibration parameters between IMU and cameras, IMU motion states, IMU biases, and gravity. To the best of our knowledge, this is the first method that performs joint gravity estimation with the IMU preintegration model.

\item The significance of higher-order IMU preintegration for discrete-time state representation in temporal calibration is highlighted, which has not been revealed previously.

\item Extensive experimental results demonstrate that our method offers unparalleled efficiency, while maintaining competitive calibration accuracy. Moreover, our method does not cause accuracy loss for VIO.

\end{itemize}

\section{Notation}

\begin{figure}[htbp]
    \centering
    \includegraphics[width=0.32\textwidth]{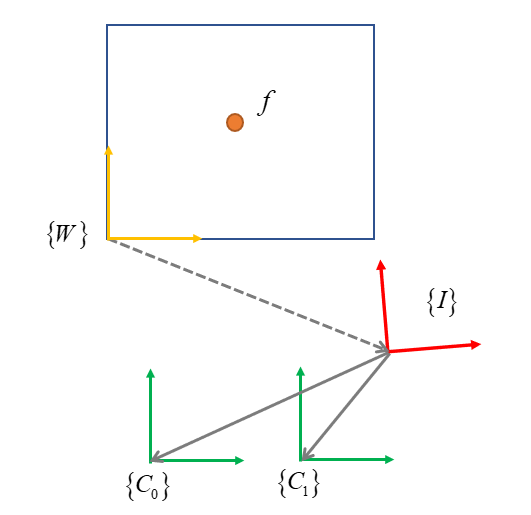}
    \caption{Coordinate frames for the IMU-Camera calibration with a calibration board.}
    \label{fig: frame}
\end{figure}

Our calibration method is illustrated with an example of spatial-temporal calibration between IMU and stereo camera. Fig. \ref{fig: frame} shows all coordinate frames involved in the calibration. $\{ W\}$ represents the world reference frame attached to the calibration board, which is static during calibration. $\{ I\} $ represents IMU coordinate frame, $\{ {C_0}\} $ and $\{ {C_1}\} $ denote left and right camera coordinate frames, respectively. $\{ I\} $, $\{ {C_0}\} $ and $\{ {C_1}\} $ are assumed to be rigidly linked.

We use ${}^W\left(  \bullet  \right)$  to represent a physical quantity in the frame $\{ W\} $. The position of a point $I$ in the frame $\{ W\}$ is expressed as ${}^W{p_I}$. The velocity of a point $I$ in the frame $\{ W\}$ is expressed as ${}^W{v_I}$. The local angular velocity of $\{ I\} $ is denoted as $\omega $. A rotation matrix is employed to represent the rotation of a rigid body. ${}_I^WR$ represents rotation from frame $\{ I\}$ to frame $\{ W\} $. ${}_I^WT$ represents 6DoF rigid body transformation from frame $\{ I\}$ to frame $\{ W\} $
\begin{equation} \label{eq: se3}
    {}_I^WT = \left[ {\begin{array}{*{20}{c}}
    {{}_I^WR}&{{}^W{p_I}}\\
    0&1
    \end{array}} \right],{}_I^WR \in SO\left( 3 \right),{}_I^WT \in SE\left( 3 \right)
\end{equation}

The 6DoF transformations between two camera frames and IMU frame are noted as spatial calibration parameters $\left\{ {{}_{{C_0}}^IT,{}_{{C_1}}^IT} \right\}$. In our formulation, the IMU time clock is treated as a time reference in the estimator. The stereo camera is assumed to be already time-synchronized, like Basalt \cite{sommer2020efficient}. This is a reasonable assumption for practical stereo-vision applications. The time offset between the camera clock and the IMU clock is the temporal calibration parameter ${t_d}$. If the image timestamp at camera clock is ${t_C}$, then the corresponding timestamp at IMU clock should be shifted with time offset
\begin{equation} \label{eq: clock}
    {t_I} = {t_C} + {t_d}
\end{equation}

The interested spatial-temporal calibration parameter set involved in this problem setting is $\left\{ {{}_{{C_0}}^IT,{}_{{C_1}}^IT,{t_d}} \right\}$. The transpose of a matrix is ${\left[  \bullet  \right]^T}$.

\section{Methodology}

Traditional IMU-Camera calibration methods typically uses continuous-time state representation (B-spline) and constructs IMU measurement model using raw measurements. Due to the high frequency of IMU measurements, continuous-time state representation leads to high-dimensional state and expensive computational cost. The main contribution of this paper is to leverage discrete-time state representation to explore how far we can accelerate the spatial-temporal calibration between IMU and cameras, without compromising accuracy. This exploration is of great significance with the vast usage of visual-inertial sensors.

Like most target-based calibration methods (Kalibr \cite{furgale2013unified}, Basalt \cite{sommer2020efficient} and MVIS \cite{yang2024multi}), we use a grid of AprilTag \cite{olson2011apriltag} as the calibration board. The coordinate frames involved in calibration are depicted in Fig. \ref{fig: frame}. The timestamp of the $i$th image is ${t_i}$. The image coordinate of the $l$th AprilTag corner ${f_l}$ detected in the $i$th image of the $n$th camera is ${}^n{u_{il}}$. Its associated 3D coordinates in $\{ W\} $, ${}^W{p_{{f_l}}}$, is known as geometric prior from the calibration board. When performing calibration, it is required to wave the sensor rig in front of a calibration board and apply sufficient motion excitation, especially in rotation \cite{mirzaei2008kalman, kelly2011visual}. The optimization variable set $\chi$ of our calibration method can be defined as a vector of several discrete-time state variables
\begin{equation} \label{eq: state}
    \begin{array}{l}
    \chi  = {\left[ {\begin{array}{*{20}{c}}
    {x_I^T}&{x_{calib}^T}
    \end{array}} \right]^T}\\
    {x_I} = {\left[ {\begin{array}{*{20}{c}}
    {x_{{I_0}}^T}& \cdots &{x_{{I_i}}^T}& \cdots &{x_{{I_M}}^T}
    \end{array}} \right]^T}\\[2pt]
    {x_{{I_i}}} = {\left[ {\begin{array}{*{20}{c}}
    {{}_{{I_i}}^W{R^T}}&{{}^Wv_{{I_i}}^T}&{{}^Wp_{{I_i}}^T}
    \end{array}} \right]^T}\\
    {x_{calib}} = {\left[ {\begin{array}{*{20}{c}}
    {{}_{{C_0}}^I{T^T}}&{{}_{{C_1}}^I{T^T}}&{{t_d}}&{b_\omega ^T}&{b_a^T}&\theta &\phi 
    \end{array}} \right]^T}
    \end{array}
\end{equation}

Where $M$ is the index of the last image. $\chi$ includes the IMU motion states at different image timestamps, ${x_I}$, as well as the calibration state, ${x_{calib}}$. IMU motion state corresponding to the $i$th image ${x_{{I_i}}}$ includes both pose and velocity $\left\{ {{}_{{I_i}}^WR,{}^W{v_{{I_i}}},{}^W{p_{{I_i}}}} \right\}$, which are expressed in world frame $\{ W\} $. Calibration state ${x_{calib}}$ contains both a set of spatial-temporal calibration parameters $\left\{ {{}_{{C_0}}^IT,{}_{{C_1}}^IT,{t_d}} \right\}$, and IMU biases $\left\{ {{b_\omega },{b_a}} \right\}$. $b_\omega$ and $b_a$ represent the gyroscope bias and the accelerometer bias, respectively. As the duration of a calibration sequence is typically short (about 1 minute, see TABLE \ref{tab: state_dim}), IMU biases can be assumed to be time-invariant, like Basalt \cite{sommer2020efficient} and iKalibr \cite{chen2025ikalibr}, as shown in TABLE \ref{tab: offline_calib}.

As gravity is included in accelerometer measurement model, and unknown in our case, it is necessary to estimate gravity ${}^Wg$, which is also expressed in world frame $\{ W\} $. The norm of gravity can be assumed to be known and remain constant\footnote{With the increasing interest and investment on the robotic exploration of extraterrestrial space \cite{delaune2021range, alberico2024structure}, IMU-Camera calibration maybe executed on other celestial body in the future, like the Moon or Mars. The gravity norm should be adapted accordingly. If gravity norm is unknown, full 3DoF gravity can be easily included in Eq. \ref{eq: state}.}, thus ${}^Wg \in {S^2}$. We use straightforward and effective spherical coordinate to parameterize gravity
\begin{equation} \label{eq: gravity}
    {}^Wg \buildrel \Delta \over = {}^Wg\left( {\rho ,\theta ,\phi } \right) = \rho \left[ {\begin{array}{*{20}{c}}
    {\cos \left( \theta  \right)\sin \left( \phi  \right)}\\
    {\sin \left( \theta  \right)\sin \left( \phi  \right)}\\
    {\cos \left( \phi  \right)}
    \end{array}} \right]
\end{equation}

By utilizing prior knowledge of gravity norm,
$\rho  = \left\| {{}^Wg} \right\| = 9.81{m \mathord{\left/ {\vphantom {m {{s^2}}}} \right. \kern-\nulldelimiterspace} {{s^2}}}$, 
the dimension of gravity to be optimized can be reduced from 3 to 2. Therefore, we only include $\left\{ {\theta ,\phi } \right\}$ in ${x_{calib}}$. For other parameterization methods of element on ${S^2}$, interested readers are referred to \cite{ling2016high, bloesch2017iterated}.

\subsection{IMU pseudo-measurement model} \label{sec: IMU}

To reduce the state dimension, multiple IMU measurements between two adjacent images are aggregated into a single pseudo-measurement, which is accomplished by IMU preintegration. On-manifold IMU preintegration is originally designed for a VIO estimator \cite{forster2016manifold}. In this section, we implement a novel on-manifold IMU preintegration to cater for the high-precision  calibration task. Our pseudo-measurement model has several differences from many existing IMU preintegration approaches \cite{forster2016manifold, usenko2019visual, delama2024equivariant, yang2024multi}:

\begin{enumerate}
   \item These existing preintegration models are formulated with Euler integration. We find that it is not sufficient to ensure that the temporal calibration accuracy is comparable to Kalibr \cite{furgale2013unified} and Basalt \cite{sommer2020efficient}. Therefore, we derive a higher-order integration (Midpoint integration) to produce more accurate IMU constraints.

   \item These existing preintegration models do not optimize gravity with the residual model (see Eq. \ref{eq: imu}), because gravity is assumed to be known at the VIO initialization stage. While, our IMU pseudo-measurement model supports gravity  optimization, addressing the unknown gravity direction with respect to the world frame built on the calibration board.
   
   \item Lastly, these existing preintegration models require different IMU biases for different IMU factors. While, our IMU pseudo-measurement model supports the constant IMU biases for all IMU factors, further reducing both state dimensions and residual dimensions.
\end{enumerate}

Next, we will detail how to build our IMU pseudo-measurement model, as well as the formulation difference with the existing preintegration model. \cite{forster2016manifold} is referred as an example. All IMU measurements from frame $i$ to frame $i + 1$ are collected for integration, with an integration interval of $\left[ {{t_i},{t_{i + 1}}} \right]$. The IMU measurements at two image timestamps, $t_i$ and $t_{i + 1}$, can be obtained by linear interpolation if necessary. The preintegration items \cite{forster2016manifold} are connected with two IMU motion states. And this connection is used to build IMU constraints. The IMU preintegration items of frame $i$ are denoted as
\begin{equation}
    {\Delta _{i,i + 1}} = \left[ {\begin{array}{*{20}{c}}
    {\Delta {R_{i,i + 1}}}\\
    {\Delta {v_{i,i + 1}}}\\
    {\Delta {p_{i,i + 1}}}
    \end{array}} \right]
\end{equation}

By equivalently modifying the equation (33) from \cite{forster2016manifold}, the calculation of preinegration items over $\left[ {{t_i},{t_{i + 1}}} \right]$ can be expressed as an iterative formulation
\begin{equation} \label{eq: euler}
    \begin{array}{l}
    \Delta {R_{i,j + 1}} = \Delta {R_{i,j}}Exp\left( {{\omega _j}\Delta t} \right)\\
    \Delta {v_{i,j + 1}} = \Delta {v_{i,j}} + \Delta {R_{i,j}}{a_j}\Delta t\\
    \Delta {p_{i,j + 1}} = \Delta {p_{i,j}} + \Delta {v_{i,j}}\Delta t + \frac{1}{2}\Delta {R_{i,j}}{a_j}\Delta {t^2}
    \end{array}
\end{equation}

Where $\Delta t$ is the time interval between two consecutive IMU measurements, $\Delta t = {t_{j + 1}} - {t_j}$. And ${t_j}$ represents IMU sampling timestamp, starting from ${t_j} = {t_i}$. During iteration, ${t_j}$ subjects to
\begin{equation} \label{eq: imu_bound}
    {t_i} \le {t_j} < {t_{j + 1}} \le {t_{i + 1}}
\end{equation}

${\omega _j}$ and ${a_j}$ in Eq. \ref{eq: euler} represent de-biased IMU angular velocity and linear acceleration measurements
\begin{equation} \label{eq: de-bias}
    \begin{array}{l}
    {\omega _j} = {{\tilde \omega }_j} - {b_\omega },
    \qquad {a_j} = {{\tilde a}_j} - {b_a}
    \end{array}
\end{equation}

${\tilde \omega _j}$ and ${\tilde a_j}$ are raw IMU angular velocity and linear acceleration measurements, respectively. The initial value of ${\Delta _{i,i + 1}}$ is set as
\begin{equation} \label{eq: ini_state}
    {\Delta _{i,i}} = \left[ {\begin{array}{*{20}{c}}
    {\Delta {R_{i,i}}}\\
    {\Delta {v_{i,i}}}\\
    {\Delta {p_{i,i}}}
    \end{array}} \right] = \left[ {\begin{array}{*{20}{c}}
    {{I_{3 \times 3}}}\\
    {{0_{3 \times 1}}}\\
    {{0_{3 \times 1}}}
    \end{array}} \right]
\end{equation}

Eq. \ref{eq: euler} is used for iterative calculation until ${t_{j + 1}} = {t_{i + 1}}$. At this time ${\Delta _{i,i + 1}}$ is determinate. Please note, this iterative process is essentially Euler integration \cite{forster2016manifold}. During the integration interval $\left[ {{t_j},{t_{j + 1}}} \right]$, only the IMU measurement at $t_j$ is used. The formulation adopted in \cite{usenko2019visual} is different from Eq. \ref{eq: euler}. For example, the rotation integration is calculated as
\begin{equation}
    \begin{array}{l}
    \Delta {R_{i,j + 1}} = \Delta {R_{i,j}}Exp\left( {{\omega _{j + 1}}\Delta t} \right)\\
    \end{array}
\end{equation}

But in this case, only the IMU measurement at $t_{j + 1}$ is used. To improve integration accuracy, an effective idea is to adopt a higher-order integration, such as Midpoint integration. Specifically, the average value of two IMU measurements is utilized to better approximate the integration. The iterative equation is improved from Eq. \ref{eq: euler} to
\begin{equation} \label{eq: midpoint}
    \begin{array}{l}
    \Delta {R_{i,j + 1}} = \Delta {R_{i,j}}Exp\left( {{{\bar \omega }_{j,j + 1}}\Delta t} \right)\\
    \Delta {v_{i,j + 1}} = \Delta {v_{i,j}} + {{\bar a}_{j,j + 1}}\Delta t\\
    \Delta {p_{i,j + 1}} = \Delta {p_{i,j}} + \Delta {v_{i,j}}\Delta t + \frac{1}{2}{{\bar a}_{j,j + 1}}\Delta {t^2}
    \end{array}
\end{equation}

Where ${\bar \omega _{j,j + 1}}$ and ${\bar a_{j,j + 1}}$ denote the average IMU angular velocity and linear acceleration measurements
\begin{equation} \label{eq: avg}
    \begin{array}{l}
    {{\bar \omega }_{j,j + 1}} = \frac{1}{2}\left( {{\omega _j} + {\omega _{j + 1}}} \right)\\
    {{\bar a}_{j,j + 1}} = \frac{1}{2}\left( {\Delta {R_{i,j}}{a_j} + \Delta {R_{i,j + 1}}{a_{j + 1}}} \right)
    \end{array}
\end{equation}

For a clear mathematical model, Eq. \ref{eq: midpoint} and Eq. \ref{eq: avg} are abstracted as a function
\begin{equation} \label{eq: f}
    {\Delta _{i,j + 1}} = f\left( {{\Delta _{i,j}},{{\tilde \omega }_j},{{\tilde \omega }_{j + 1}},{{\tilde a}_j},{{\tilde a}_{j + 1}},{b_\omega },{b_a}} \right)
\end{equation}

Using the simpler form above, the Jacobian of ${\Delta _{i,j + 1}}$ with respect to IMU biases can be clearly expressed with a recursive formulation
\begin{equation} \label{eq: jacobian_bias}
    \begin{array}{l}
    \frac{{\partial {\Delta _{i,j + 1}}}}{{\partial {b_\omega }}} = \frac{{\partial f}}{{\partial {\Delta _{i,j}}}}\frac{{\partial {\Delta _{i,j}}}}{{\partial {b_\omega }}} + \frac{{\partial f}}{{\partial {b_\omega }}}\\
    \frac{{\partial {\Delta _{i,j + 1}}}}{{\partial {b_a}}} = \frac{{\partial f}}{{\partial {\Delta _{i,j}}}}\frac{{\partial {\Delta _{i,j}}}}{{\partial {b_a}}} + \frac{{\partial f}}{{\partial {b_a}}}
    \end{array}
\end{equation}

Please note, the index in the above equation runs from $j$ to $j+1$. The corresponding initial Jacobian is set as
\begin{equation} \label{eq: ini_bias_jacobian}
    \begin{array}{l}
    \frac{{\partial {\Delta _{i,i}}}}{{\partial {b_\omega }}} = {0_{9 \times 3}}, \qquad
    \frac{{\partial {\Delta _{i,i}}}}{{\partial {b_a}}} = {0_{9 \times 3}}
    \end{array}
\end{equation}

We adopt left perturbation to calculate Jacobian for the element on Lie group \cite{sola2018micro}. With the iteration of Eq. \ref{eq: midpoint}, IMU noise generated from $\left\{ {{\Delta _{i,j}},{{\tilde \omega }_j},{{\tilde \omega }_{j + 1}},{{\tilde a}_j},{{\tilde a}_{j + 1}}} \right\}$ (see Eq. \ref{eq: f}) is propagated as
\begin{equation} \label{eq: jacobian_cov}
    \begin{array}{l}
    {\Sigma _{i,j + 1}} = \left( {\frac{{\partial f}}{{\partial {\Delta _{i,j}}}}} \right){\Sigma _{i,j}}{\left( {\frac{{\partial f}}{{\partial {\Delta _{i,j}}}}} \right)^T}\\
     + \left( {\frac{{\partial f}}{{\partial {{\tilde \omega }_j}}}} \right){\Sigma _\omega }{\left( {\frac{{\partial f}}{{\partial {{\tilde \omega }_j}}}} \right)^T} + \left( {\frac{{\partial f}}{{\partial {{\tilde \omega }_{j + 1}}}}} \right){\Sigma _\omega }{\left( {\frac{{\partial f}}{{\partial {{\tilde \omega }_{j + 1}}}}} \right)^T}\\
     + \left( {\frac{{\partial f}}{{\partial {{\tilde a}_j}}}} \right){\Sigma _a}{\left( {\frac{{\partial f}}{{\partial {{\tilde a}_j}}}} \right)^T} + \left( {\frac{{\partial f}}{{\partial {{\tilde a}_{j + 1}}}}} \right){\Sigma _a}{\left( {\frac{{\partial f}}{{\partial {{\tilde a}_{j + 1}}}}} \right)^T}
    \end{array}
\end{equation}

Where ${\Sigma _\omega }$ and ${\Sigma _a}$ represent noise covariance for IMU angular velocity and linear acceleration measurements. Given this iterative formulation, the covariance of ${\Delta _{i,i + 1}}$, ${\Sigma _{i,i + 1}}$, is obtained together when the integration of ${\Delta _{i,i + 1}}$ is finished. The initial value of ${\Sigma _{i,i + 1}}$ is set as
\begin{equation} \label{eq: ini_cov}
    {\Sigma _{i,i}} = {0_{9 \times 9}}
\end{equation}

Due to page limit, the analytical on-manifold Jacobian of $f$ (Eq. \ref{eq: f}) with respect to all involved variables, $\left\{ {{\Delta _{i,j}},{{\tilde \omega }_j},{{\tilde \omega }_{j + 1}},{{\tilde a}_j},{{\tilde a}_{j + 1}},{b_\omega },{b_a}} \right\}$, are omitted here and can be found in \cite[Sec. 6.10]{song2025calibration}. Given these derivatives, Eq. \ref{eq: jacobian_bias} and Eq. \ref{eq: jacobian_cov} become computable now.

The relationship between ${\Delta _{i,i + 1}}$ and two consecutive IMU states (${x_{{I_i}}}$ and ${x_{{I_{i + 1}}}}$, see Eq. \ref{eq: state}) models motion constraint via IMU integration. The corresponding IMU residuals\footnote{We do not incorporate IMU biases update described in Section VI.C of \cite{forster2016manifold}. Instead, we perform reintegration for each iteration (see Fig. \ref{fig: state shift}). Moreover, residuals of IMU biases (Section VI.E of \cite{forster2016manifold}) are removed.} are constructed as
\begin{equation} \label{eq: imu}
    \begin{array}{l}
    {r_{\Delta {R_{i,i + 1}}}} = Log\left( {\Delta {R_{i,i + 1}}{}_{{I_{i + 1}}}^W{R^T}{}_{{I_i}}^WR} \right)\\
    {r_{\Delta {v_{i,i + 1}}}} = {}_{{I_i}}^W{R^T}\left( {{}^W{v_{{I_{i + 1}}}} - {}^W{v_{{I_i}}} - {}^Wgdt} \right) - \Delta {v_{i,i + 1}}\\
    {r_{\Delta {p_{i,i + 1}}}} = {}_{{I_i}}^W{R^T}\left( {{}^W{p_{{I_{i + 1}}}} - {}^W{p_{{I_i}}} - {}^W{v_{{I_i}}}dt - \frac{1}{2}{}^Wgd{t^2}} \right)\\
    \qquad\qquad\quad - \Delta {p_{i,i + 1}}\\
    {r_{{I_{i,i + 1}}}} \buildrel \Delta \over = {r_{{I_{i,i + 1}}}}\left( {{x_{{I_i}}},{x_{{I_{i + 1}}}},{b_\omega },{b_a},\theta ,\phi } \right) = \left[ {\begin{array}{*{20}{c}}
    {{r_{\Delta {R_{i,i + 1}}}}}\\
    {{r_{\Delta {v_{i,i + 1}}}}}\\
    {{r_{\Delta {p_{i,i + 1}}}}}
    \end{array}} \right]
    \end{array}
\end{equation}

Where $Log \left(  \bullet  \right)$ maps the element on a Lie group to the tangent space vector \cite{sola2018micro}. $dt$ is the time gap between two images, $dt = {t_{i + 1}} - {t_i}$. For the Jacobian of residual ${r_{{I_{i,i + 1}}}}$ with respect to ${x_{{I_i}}}$ and ${x_{{I_{i + 1}}}}$, we adopt left perturbation for the element on Lie group, as previous derivatives of $f$ (Eq. \ref{eq: f}). The Jacobian of ${r_{{I_{i,i + 1}}}}$ with respect to IMU biases, $\left\{ {{b_\omega },{b_a}} \right\}$, can be obtained from Eq. \ref{eq: jacobian_bias}. Finally, the Jacobian of ${r_{{I_{i,i + 1}}}}$ with respect to the gravity direction $\left\{ {\theta ,\phi } \right\}$ can be computed according to the chain rule
\begin{equation}
    \begin{array}{l}
    {g_s} = {\left[ {\begin{array}{*{20}{c}}
    \theta &\phi 
    \end{array}} \right]^T}\\
    \frac{{\partial {r_{{I_{i,i + 1}}}}}}{{\partial {g_s}}} = \frac{{\partial {r_{{I_{i,i + 1}}}}}}{{\partial {}^Wg}}\frac{{\partial {}^Wg}}{{\partial {g_s}}}
    \end{array}
\end{equation}

Now, we complete the mathematical details of the IMU pseudo-measurement model between two adjacent frames.

\subsection{Camera measurement model with time offset} \label{sec: Cam}

The IMU pseudo-measurement model described in the previous section can only constrain IMU states, IMU biases, and gravity (see Eq. \ref{eq: state} and Eq. \ref{eq: imu}). And it is unable to link to interested spatial-temporal calibration parameters. To this end, we introduce a camera measurement model with time offset, to provide additional constraints on both the spatial and temporal calibration parameters
\begin{equation} \label{eq: cam}
    {}^n{r_{il}} = \pi \left( {{}_{{C_n}}^I{T^{ - 1}}{}_{{I_i}}^WT{{\left( {{t_i} + {t_d}} \right)}^{ - 1}}{}^W{p_{{f_l}}}} \right) - {}^n{u_{il}}
\end{equation}

$\pi \left(  \bullet  \right)$ is a fixed camera projection function \cite{usenko2018double}. The camera intrinsic parameters are assumed to be pre-calibrated. ${}^n{r_{il}}$ represents the pixel residual (reprojection error) obtained by the $n$th camera, observing the $l$th AprilTag corner at the $i$th frame. The measurement covariance is typically set by engineering experience, such as fixed at 1 pixel. The analytical on-manifold Jacobian of ${}^n{r_{il}}$ with respect to all involved variables can be found in \cite[Sec. 6.11]{song2025calibration}.

\subsection{Full-batch nonlinear least squares optimization} \label{sec: optimization}


By integrating all raw image pixel measurements and IMU pseudo-measurements, we formulate the full-batch nonlinear least squares optimization as
\begin{equation} \label{eq: opt}
    \begin{split}
    \begin{array}{l}
    \chi  = \arg \min \left\{ \sum\limits_{n = 0}^1 \sum\limits_{i = 0}^M {\sum\limits_{l \in {K_{ni}}} {\rho \left( {\left\| {{}^n{r_{il}}} \right\|_{{\Sigma _C}}^2} \right)} } \right. \\
    \left. \qquad\qquad\qquad + \sum\limits_{i = 0}^{M - 1} {\left\| {{r_{{I_{i,i + 1}}}}} \right\|_{{\Sigma _{i,i + 1}}}^2} \right\}\\
    {}^n{r_{il}} = \pi \left( {{}_{{C_n}}^I{T^{ - 1}}{}_{{I_i}}^WT{{\left( {{t_i} + {t_d}} \right)}^{ - 1}}{}^W{p_{{f_l}}}} \right) - {}^n{u_{il}}\\
    {r_{{I_{i,i + 1}}}} \buildrel \Delta \over = {r_{{I_{i,i + 1}}}}\left( {{x_{{I_i}}},{x_{{I_{i + 1}}}},{b_\omega },{b_a},\theta ,\phi } \right)
    \end{array}
    \end{split}
\end{equation}

Where $M$ is the index of the last image. ${}^n{r_{il}}$ is the pixel residual from the $n$th camera. ${r_{{I_{i,i + 1}}}}$ is the IMU pseudo-measurement residual. Detailed definitions of these two types of residuals are provided in Section \ref{sec: IMU} and Section \ref{sec: Cam}. ${K_{ni}}$ represent the set of corner points observed by the $n$th camera at the $i$th frame. $\rho \left(  \bullet  \right)$ is a robust kernel function \cite{chebrolu2021adaptive}. Typically, robust Huber kernel function is used to mitigate the impact of pixel observation outliers. We refer to Basalt \cite{sommer2020efficient} to provide the initial guess for the optimization variable set $\chi$. Levenberg-Marquardt algorithm is adopted to minimize Eq. \ref{eq: opt} and update the optimal estimation iteratively.

At the end of each iteration, the timestamps of all images are shifted with the current estimate of time offset
\begin{equation}
    {t_i} \leftarrow {t_i} + {t_d}
\end{equation}

The new integration interval for each IMU pseudo-measurement at the next iteration should be updated as
\begin{equation}
    \left[ {{t_i},{t_{i + 1}}} \right] \leftarrow \left[ {\left( {{t_i} + {t_d}} \right),\left( {{t_{i + 1}} + {t_d}} \right)} \right]
\end{equation}

\begin{figure}[htbp]
    \centering
    \includegraphics[width=0.42\textwidth]{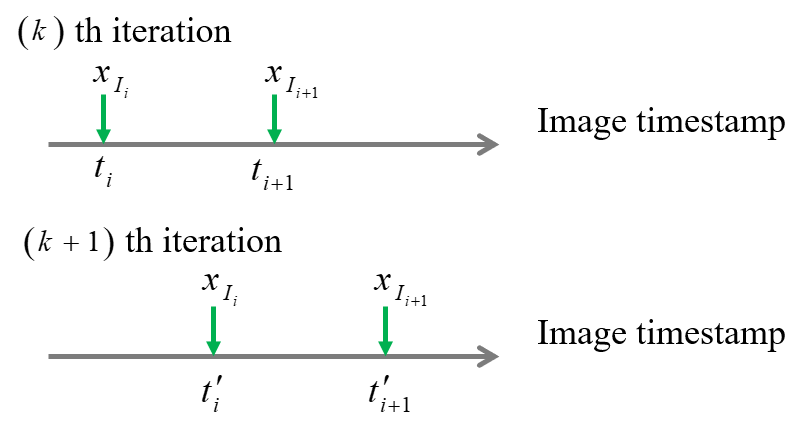}
    \caption{Time shift of each IMU motion state corresponding to image. After the time shift of images, ${t_i}$ and ${t_{i + 1}}$ become ${t'_i}$ and ${t'_{i + 1}}$, respectively. ${t'_i} = {t_i} + {t_d}$, ${t'_{i + 1}} = {t_{i + 1}} + {t_d}$.}
    \label{fig: state shift}
\end{figure}

Since IMU motion states (${x_I}$ in Eq. \ref{eq: state}) are dependent on image time, they are shifted together in time domain with the update of time offset, as depicted in Fig. \ref{fig: state shift}.

\section{Results} \label{sec: results}

We benchmark the calibration accuracy and computational efficiency of the proposed method with two state-of-the-art (SOTA) baseline methods, \textbf{Kalibr} \cite{furgale2013unified} and \textbf{Basalt} \cite{sommer2020efficient}, both of which adopt continuous-time state representation. Compared to Kalibr, Basalt accelerates calibration through split B-spline representation and advanced derivative calculation method. Default optimized configurations for them are used for fair comparison, like the experiments in MVIS \cite{yang2024multi}. To demonstrate the performance of our method, experiments are designed to address the following questions

\begin{enumerate}
   \item Is the spatial-temporal calibration accuracy of the proposed method comparable to Kalibr and Basalt?
   \item Is the optimization efficiency of the proposed method much faster than Kalibr and Basalt?
   \item As illustrated in Section \ref{sec: IMU}, do we need to replace Euler integration with Midpoint integration?
   \item Does VIO incur accuracy loss by using the proposed calibration method?
\end{enumerate}

We perform experiments with IMU-Camera calibration sequences from two popular VIO datasets, EuRoC \cite{burri2016euroc}, and TUM-VI \cite{schubert2018tum}. These two datasets provide stereo image and IMU data at 20Hz and 200Hz, respectively. To demonstrate the calibration performance at different camera frequencies, new calibration sequences are generated by reducing the original image frequency from 20Hz to 10Hz, and 5Hz.

The reference value of the temporal calibration parameter can be obtained from the dataset providers. It is convenient to reset the desired time offset by manually shifting the timestamp of IMU data with a certain value. The shifted time offset modifies the new reference value of the temporal calibration parameter. The reference value of the spatial calibration parameter is obtained by calibrating the original sequence with Kalibr.

The proposed calibration method is evaluated with two variants with different IMU integration methods for the IMU pseudo-measurement model (Section \ref{sec: IMU}). If the IMU pseudo-measurement model uses Euler integration, our method is referred to as \textbf{"Ours (Euler)"}. Specifically, the integration model in \cite{usenko2019visual} is used. If Midpoint integration (see Eq. \ref{eq: avg} and Eq. \ref{eq: midpoint}) is employed, our method is referred to as \textbf{"Ours (Midpoint)"}. All the experiments are conducted on a laptop computer with an Intel(R) Xeon(R) W-10855M CPU @ 2.80GHz, and 16 GB of RAM.

\subsection{EuRoC dataset}

\begin{table*}
  \caption{Average metrics of different calibration methods on the EuRoC dataset. Evaluation metrics include the average RMSE results of spatial-temporal calibration (rotation, translation, time offset), reprojection error, optimization time and speed up of our method compared to SOTA baselines.}
  \centering
  \aboverulesep=0ex
  \belowrulesep=0ex
  \scalebox{1.05}{
  \begin{tabular}{@{}c|c|cccccc@{}}
    \toprule
    \multirow{2}{*}{Metrics (unit)} & \multirow{2}{*}{Methods} & \multicolumn{2}{c}{20 Hz} & \multicolumn{2}{c}{10 Hz} & \multicolumn{2}{c}{5 Hz} \\ 
    \cmidrule(lr){3-4} \cmidrule(lr){5-6} \cmidrule(lr){7-8} & & {Camera $C_0$} & {Camera $C_1$} & {Camera $C_0$} & {Camera $C_1$} & {Camera $C_0$} & {Camera $C_1$} \\
    \midrule
    \multirow{4}{*}{Rotation (degree)} & Kalibr & 0.000 $\pm$ 0.000 & 0.000 $\pm$ 0.000  & 0.009 $\pm$ 0.000 & 0.009 $\pm$ 0.000 & 0.024 $\pm$ 0.000 & 0.024 $\pm$ 0.000  \\
    & Basalt & 0.497 $\pm$ 0.622 & 0.502 $\pm$ 0.621 & 0.486 $\pm$ 0.621 & 0.492 $\pm$ 0.620 & 0.437 $\pm$ 0.649 & 0.444 $\pm$ 0.648   \\
    & Ours (Euler) & 0.012 $\pm$ 0.000 & 0.023 $\pm$ 0.000 & 0.014 $\pm$ 0.000 & 0.028 $\pm$ 0.000 & 0.090 $\pm$ 0.000 & 0.103 $\pm$ 0.000    \\
    & Ours (Midpoint) & 0.015 $\pm$ 0.000 & 0.014 $\pm$ 0.000 & 0.009 $\pm$ 0.000 & 0.015 $\pm$ 0.000 & 0.041 $\pm$ 0.000 & 0.047 $\pm$ 0.000   \\
    \hline
    \multirow{4}{*}{Translation (cm)} & Kalibr & 0.000 $\pm$ 0.000 & 0.000 $\pm$ 0.000 & 0.025 $\pm$ 0.000 & 0.025 $\pm$ 0.000 & 0.078 $\pm$ 0.000 & 0.078 $\pm$ 0.000  \\
    & Basalt & 0.870 $\pm$ 1.117 & 0.870 $\pm$ 1.116 & 0.798 $\pm$ 1.054 & 0.798 $\pm$ 1.052 & 0.506 $\pm$ 0.604 & 0.506 $\pm$ 0.602   \\
    & Ours (Euler) & 0.045 $\pm$ 0.000 & 0.049 $\pm$ 0.000 & 0.073 $\pm$ 0.000 & 0.075 $\pm$ 0.000 & 0.301 $\pm$ 0.000 & 0.296 $\pm$ 0.000    \\
    & Ours (Midpoint) & 0.039 $\pm$ 0.000 & 0.048 $\pm$ 0.000 & 0.039 $\pm$ 0.000 & 0.050 $\pm$ 0.000 & 0.047 $\pm$ 0.000 & 0.058 $\pm$ 0.000  \\
    \hline
    \multirow{4}{*}{Time offset (ms)} & Kalibr & \multicolumn{2}{c}{0.035 $\pm$ 0.000} & \multicolumn{2}{c}{0.044 $\pm$ 0.000}  & \multicolumn{2}{c}{0.066 $\pm$ 0.000}   \\
    & Basalt & \multicolumn{2}{c}{12.971 $\pm$ 17.839} & \multicolumn{2}{c}{11.397 $\pm$ 15.590}  & \multicolumn{2}{c}{10.529 $\pm$ 16.550}    \\
    & Ours (Euler) & \multicolumn{2}{c}{2.456 $\pm$ 0.000} & \multicolumn{2}{c}{2.449 $\pm$ 0.000}  & \multicolumn{2}{c}{2.483 $\pm$ 0.000}    \\
    & Ours (Midpoint) & \multicolumn{2}{c}{0.043 $\pm$ 0.000} & \multicolumn{2}{c}{0.068 $\pm$ 0.000}  & \multicolumn{2}{c}{0.158 $\pm$ 0.000}   \\
    \hline
    \hline
    \multirow{4}{*}{\makecell{Reprojection error\\ (pixel)}} & Kalibr & \multicolumn{2}{c}{0.373 $\pm$ 0.000} & \multicolumn{2}{c}{0.375 $\pm$ 0.000}  & \multicolumn{2}{c}{0.374 $\pm$ 0.000}   \\
    & Basalt & \multicolumn{2}{c}{0.263 $\pm$ 0.071} & \multicolumn{2}{c}{0.266 $\pm$ 0.072}  & \multicolumn{2}{c}{0.246 $\pm$ 0.060}    \\
    & Ours (Euler) & \multicolumn{2}{c}{0.208 $\pm$ 0.000} & \multicolumn{2}{c}{0.210 $\pm$ 0.000}  & \multicolumn{2}{c}{0.217 $\pm$ 0.000}    \\
    & Ours (Midpoint) & \multicolumn{2}{c}{0.209 $\pm$ 0.000} & \multicolumn{2}{c}{0.211 $\pm$ 0.000}  & \multicolumn{2}{c}{0.213 $\pm$ 0.000}   \\
    \hline
    \multirow{4}{*}{\makecell{Optimization time\\ (s)}} & Kalibr & \multicolumn{2}{c}{144.170 $\pm$ 0.837} & \multicolumn{2}{c}{108.134 $\pm$ 0.496}  & \multicolumn{2}{c}{51.255 $\pm$ 0.419}   \\
    & Basalt & \multicolumn{2}{c}{14.919 $\pm$ 2.495} & \multicolumn{2}{c}{15.064 $\pm$ 2.284}  & \multicolumn{2}{c}{15.730 $\pm$ 3.027}    \\
    & Ours (Euler) & \multicolumn{2}{c}{0.289 $\pm$ 0.004} & \multicolumn{2}{c}{0.136 $\pm$ 0.015}  & \multicolumn{2}{c}{0.189 $\pm$ 0.338}    \\
    & Ours (Midpoint) & \multicolumn{2}{c}{0.290 $\pm$ 0.004} & \multicolumn{2}{c}{0.140 $\pm$ 0.012}  & \multicolumn{2}{c}{0.081 $\pm$ 0.001}   \\
    \hline
    \multicolumn{2}{c|}{\makecell{Speedup of Ours (Midpoint)\\ compared to Kalibr}} & \multicolumn{2}{c}{497.138x} & \multicolumn{2}{c}{772.386x}  & \multicolumn{2}{c}{632.778x}   \\
    \hline
    \multicolumn{2}{c|}{\makecell{Speedup of Ours (Midpoint)\\ compared to Basalt}} & \multicolumn{2}{c}{51.445x} & \multicolumn{2}{c}{107.600x}  & \multicolumn{2}{c}{194.198x}   \\
    \bottomrule
  \end{tabular}}
  \label{tab: euroc_calib}
\end{table*}

\begin{table*}
  \caption{Average metrics of different calibration methods on the TUM-VI dataset. Evaluation metrics include the average RMSE results of spatial-temporal calibration (rotation, translation, time offset), reprojection error, optimization time and speed up of our method compared to SOTA baselines.}
  \centering
  \aboverulesep=0ex
  \belowrulesep=0ex
  \scalebox{1.05}{
  \begin{tabular}{@{}c|c|cccccc@{}}
    \toprule
    \multirow{2}{*}{Metrics (unit)} & \multirow{2}{*}{Methods} & \multicolumn{2}{c}{20 Hz} & \multicolumn{2}{c}{10 Hz} & \multicolumn{2}{c}{5 Hz} \\ 
    \cmidrule(lr){3-4} \cmidrule(lr){5-6} \cmidrule(lr){7-8} & & {Camera $C_0$} & {Camera $C_1$} & {Camera $C_0$} & {Camera $C_1$} & {Camera $C_0$} & {Camera $C_1$} \\
    \midrule
    \multirow{4}{*}{Rotation (degree)} & Kalibr & 0.000 $\pm$ 0.000 & 0.000 $\pm$ 0.000 & 0.001 $\pm$ 0.000 & 0.001 $\pm$ 0.000 & 0.002 $\pm$ 0.000 & 0.002 $\pm$ 0.000  \\
    & Basalt & 0.002 $\pm$ 0.000 & 0.003 $\pm$ 0.000 & 0.002 $\pm$ 0.000 & 0.003 $\pm$ 0.000 & 0.002 $\pm$ 0.000 & 0.003 $\pm$ 0.000  \\
    & Ours (Euler) & 0.002 $\pm$ 0.000 & 0.003 $\pm$ 0.000 & 0.003 $\pm$ 0.000 & 0.002 $\pm$ 0.000 & 0.004 $\pm$ 0.000 & 0.002 $\pm$ 0.000    \\
    & Ours (Midpoint) & 0.003 $\pm$ 0.000 & 0.002 $\pm$ 0.000 & 0.004 $\pm$ 0.000 & 0.001$\pm$ 0.000  & 0.005 $\pm$ 0.000 & 0.001 $\pm$ 0.000   \\
    \hline
    \multirow{4}{*}{Translation (cm)} & Kalibr & 0.000 $\pm$ 0.000 & 0.000 $\pm$ 0.000 & 0.007 $\pm$ 0.000 & 0.007 $\pm$ 0.000 & 0.013 $\pm$ 0.000 & 0.013 $\pm$ 0.000 \\
    & Basalt & 0.069 $\pm$ 0.000 & 0.077 $\pm$ 0.000 & 0.069 $\pm$ 0.000 & 0.077 $\pm$ 0.000 & 0.069 $\pm$ 0.000 & 0.077 $\pm$ 0.000   \\
    & Ours (Euler) & 0.011 $\pm$ 0.000 & 0.007 $\pm$ 0.000 & 0.015 $\pm$ 0.000 & 0.009 $\pm$ 0.000 & 0.020 $\pm$ 0.000 & 0.014 $\pm$ 0.000   \\
    & Ours (Midpoint) & 0.017 $\pm$ 0.000 & 0.014 $\pm$ 0.000 & 0.021 $\pm$ 0.000 & 0.017 $\pm$ 0.000 & 0.027 $\pm$ 0.000 & 0.022 $\pm$ 0.000  \\
    \hline
    \multirow{4}{*}{Time offset (ms)} & Kalibr & \multicolumn{2}{c}{0.165 $\pm$ 0.000} & \multicolumn{2}{c}{0.163 $\pm$ 0.000}  & \multicolumn{2}{c}{0.167 $\pm$ 0.000}   \\
    & Basalt & \multicolumn{2}{c}{0.168 $\pm$ 0.000} & \multicolumn{2}{c}{0.168 $\pm$ 0.000}  & \multicolumn{2}{c}{0.168 $\pm$ 0.000}    \\
    & Ours (Euler) & \multicolumn{2}{c}{2.341 $\pm$ 0.000} & \multicolumn{2}{c}{2.341 $\pm$ 0.000}  & \multicolumn{2}{c}{2.339 $\pm$ 0.000}    \\
    & Ours (Midpoint) & \multicolumn{2}{c}{0.165 $\pm$ 0.000} & \multicolumn{2}{c}{0.164 $\pm$ 0.000}  & \multicolumn{2}{c}{0.165 $\pm$ 0.000}   \\
    \hline
    \hline
    \multirow{4}{*}{\makecell{Reprojection error\\ (pixel)}} & Kalibr & \multicolumn{2}{c}{0.085 $\pm$ 0.000} & \multicolumn{2}{c}{0.085 $\pm$ 0.000}  & \multicolumn{2}{c}{0.085 $\pm$ 0.000}   \\
    & Basalt & \multicolumn{2}{c}{0.087 $\pm$ 0.000} & \multicolumn{2}{c}{0.087 $\pm$ 0.000}  & \multicolumn{2}{c}{0.087 $\pm$ 0.000}    \\
    & Ours (Euler) & \multicolumn{2}{c}{0.087 $\pm$ 0.000} & \multicolumn{2}{c}{0.087 $\pm$ 0.000}  & \multicolumn{2}{c}{0.087 $\pm$ 0.000}    \\
    & Ours (Midpoint) & \multicolumn{2}{c}{0.087 $\pm$ 0.000} & \multicolumn{2}{c}{0.088 $\pm$ 0.000}  & \multicolumn{2}{c}{0.088 $\pm$ 0.000}   \\
    \hline
    \multirow{4}{*}{\makecell{Optimization time\\ (s)}} & Kalibr & \multicolumn{2}{c}{100.420 $\pm$ 5.842} & \multicolumn{2}{c}{63.454 $\pm$ 4.071}  & \multicolumn{2}{c}{44.640 $\pm$ 2.465}   \\
    & Basalt & \multicolumn{2}{c}{5.890 $\pm$ 0.651} & \multicolumn{2}{c}{5.901 $\pm$ 0.686}  & \multicolumn{2}{c}{5.959 $\pm$ 0.899}    \\
    & Ours (Euler) & \multicolumn{2}{c}{0.201 $\pm$ 0.013} & \multicolumn{2}{c}{0.093 $\pm$ 0.011}  & \multicolumn{2}{c}{0.049 $\pm$ 0.005}    \\
    & Ours (Midpoint) & \multicolumn{2}{c}{0.196 $\pm$ 0.020} & \multicolumn{2}{c}{0.094 $\pm$ 0.011}  & \multicolumn{2}{c}{0.050 $\pm$ 0.005}   \\
    \hline
    \multicolumn{2}{c|}{\makecell{Speedup of Ours (Midpoint)\\ compared to Kalibr}} & \multicolumn{2}{c}{512.347x} & \multicolumn{2}{c}{675.043x}  & \multicolumn{2}{c}{892.800x}   \\
    \hline
    \multicolumn{2}{c|}{\makecell{Speedup of Ours (Midpoint)\\ compared to Basalt}} & \multicolumn{2}{c}{30.051x} & \multicolumn{2}{c}{62.777x}  & \multicolumn{2}{c}{119.180x}   \\
    \bottomrule
  \end{tabular}}
  \label{tab: tum_calib}
\end{table*}

For each sequence with specific image frequency, we shift the timestamp of IMU data from -50ms to 50ms, with 10ms increment. In this way, the number of calibration sequences for each image frequency is increased to 11. The average RMSE results (rotation, translation, time offset) of spatial-temporal calibration with different methods are summarized under different image frequencies in TABLE \ref{tab: euroc_calib}. When the image frequency is 20Hz, the rotation and translation RMSE of Kalibr is 0 degree and 0 cm, respectively. This is because the reference value of spatial calibration parameter is obtained via Kalibr itself. The time offset RMSE of Kalibr is only 0.035 ms. With the decrease of image frequency, Kalibr's spatial-temporal calibration results gradually deviate from the reference value. This deviation is very small. Specifically, the rotation deviation is less than 0.05 degree, the translation deviation is less than 0.1 cm, and the time offset deviation is less than 0.1 ms. These results demonstrate Kalibr's excellent and reliable calibration accuracy.

Another baseline, Basalt, exhibits large RMSE results for spatial-temporal calibration. Particularly, the time offset RMSE of Basalt is greater than 10 ms for all image frequencies. The large calibration error indicates that Basalt performs unreliable calibration, although its reprojection error is smaller than Kalibr. By carefully examining the implementation of Kalibr and Basalt, we found that the poor performance of Basalt can be attributed to its coarse initialization for time offset (set as 0 ms).

The RMSE results of Ours (Euler) are better than Basalt, and validate that our method is more robust to different time offsets. We note that the time offset RMSE (around 2.5 ms) of Ours (Euler) is greater than Kalibr, even though it is already less than the measurement period of IMU (5 ms). Time offset estimation issue for discrete-time state representation is also reported in a recent SLAM benchmark study \cite{cioffi2022continuous}. Another variation of our method, Ours (Midpoint), successfully addresses this issue by reducing the time offset RMSE to less than 0.2 ms. In addition, the rotation RMSE of Ours (Midpoint) is less than 0.05 degree, and the translation RMSE of Ours (Midpoint) is less than 0.1 cm. Impressive numerical results demonstrate that the calibration accuracy of Ours (Midpoint) is comparable to Kalibr, and the necessity of upgrading IMU integration from Euler integration to Midpoint integration (Section \ref{sec: IMU}).

\subsubsection{Advantages on efficiency}

Apart from the accuracy metric, TABLE \ref{tab: euroc_calib} also shows the efficiency metric, more specifically, the optimization time of each calibration method. When the image frequency is 20Hz, the average optimization time of Kalibr is 144.17 s, which is approximately twice the duration of the calibration sequence (71.9 s). Kalibr decreases the optimization time by reducing the image frequency, because camera measurements become less. The optimization time of Basalt can be reduced to around 15 s. While Ours (Euler) is much faster than Kalibr and Basalt, reducing the optimization time to less than 0.3 s. Our method significantly benefits from the lower state dimensions (see TABLE \ref{tab: state_dim}), and lower residual dimensions (Section \ref{sec: IMU}). When the image frequency is 5Hz, the optimization time of Ours (Midpoint) is even faster than Ours (Euler), reducing to 0.081 s. This is achieved by faster convergence. On average, Ours (Midpoint) is 634x faster than Kalibr.

\subsubsection{Impact on VIO accuracy}

To evaluate the influence of different calibration methods on the localization accuracy of VIO, a SOTA VIO estimator, Open-VINS \cite{geneva2020openvins}, is tested for localization experiments. Spatial-temporal parameters are obtained with the original calibration sequence. Open-VINS has the capability to perform online spatial-temporal calibration. If online calibration is enabled, it is denoted as Open-VINS (w. calib). Otherwise, it is denoted as Open-VINS (wo. calib). Absolute trajectory error (ATE) \cite{zhang2018tutorial} is utilized as the accuracy metric, and obtained by aligning the estimated trajectory with the groundtruth trajectory in posyaw mode. Results are reported in TABLE \ref{tab: vio_euroc}. This table demonstrates that using our calibration method does not cause accuracy loss for VIO, compared to Kalibr or Basalt.

\begin{table}
\caption{ATE (meter) Comparison on the EuRoC Dataset with different calibration parameters from Kalibr, Basalt and Ours (Midpoint), respectively.}
\centering
\label{tab: vio_euroc}
\begin{center}
\scalebox{0.95}{
\begin{tabular}{|c|ccc|ccc|}
\hline
\multirow{2}{*}{\makecell{Sequence}} & \multicolumn{3}{c|}{Open-VINS (w. calib)} & \multicolumn{3}{c|}{Open-VINS (wo. calib)} \\
\cline{2-7}   & Kalibr & Basalt & Ours & Kalibr & Basalt & Ours \\
\hline
MH\_01\_easy & 0.056 & 0.053 & 0.053 & 0.035 & 0.036 & 0.042 \\
MH\_02\_easy & 0.058 & 0.057 & 0.062 & 0.046 & 0.046 & 0.052 \\
MH\_03\_medium & 0.067 & 0.056 & 0.052 & 0.055 & 0.056 & 0.050 \\
MH\_04\_difficult & 0.068 & 0.076 & 0.069 & 0.056 & 0.053 & 0.048 \\
MH\_05\_difficult & 0.051 & 0.047 & 0.055 & 0.045 & 0.045 & 0.047 \\
V1\_01\_easy & 0.106 & 0.088 & 0.124 & 0.133 & 0.130 & 0.097 \\
V1\_02\_medium & 0.105 & 0.071 & 0.072 & 0.072 & 0.079 & 0.065 \\
V1\_03\_difficult & 0.110 & 0.183 & 0.168 & 0.127 & 0.131 & 0.103 \\
V2\_01\_easy & 0.127 & 0.159 & 0.139 & 0.123 & 0.111 & 0.102 \\
V2\_02\_medium & 0.370 & 0.170 & 0.220 & 0.182 & 0.225 & 0.163 \\
V2\_03\_difficult & 0.364 & 0.271 & 0.271 & 0.226 & 0.195 & 0.202 \\
\hline\hline
Avg & 0.135 & 0.112 & 0.117 & 0.100 & 0.101 & 0.088 \\
\hline
\end{tabular}}
\end{center}
\end{table}

\begin{table}
\caption{ATE (meter) Comparison on the TUM-VI Dataset with different calibration parameters from Kalibr, Basalt and Ours (Midpoint), respectively.}
\centering
\label{tab: vio_tumvi}
\begin{center}
\begin{tabular}{|c|ccc|ccc|}
\hline
\multirow{2}{*}{\makecell{Sequence}} & \multicolumn{3}{c|}{Open-VINS (w. calib)} & \multicolumn{3}{c|}{Open-VINS (wo. calib)} \\
\cline{2-7}   & Kalibr & Basalt & Ours & Kalibr & Basalt & Ours \\
\hline
room1 & 0.058 & 0.064 & 0.054 & 0.052 & 0.066 & 0.058 \\
room2 & 0.090 & 0.102 & 0.104 & 0.055 & 0.063 & 0.056 \\
room3 & 0.083 & 0.083 & 0.076 & 0.092 & 0.071 & 0.077 \\
room4 & 0.030 & 0.033 & 0.042 & 0.036 & 0.032 & 0.032 \\
room5 & 0.089 & 0.088 & 0.095 & 0.109 & 0.092 & 0.090 \\
\hline\hline
Avg & 0.070 & 0.074 & 0.074 & 0.069 & 0.065 & 0.063 \\
\hline
\end{tabular}
\end{center}
\end{table}

\subsection{TUM-VI dataset}

The collection platform of the TUM-VI dataset is shown in Fig. \ref{fig: vi-calib}. Like the experiments for EuRoC dataset, we perform augmentation operation for calibration sequence by shifting IMU timestamp manually. The average RMSE results of spatial-temporal calibration with different methods are presented in TABLE \ref{tab: tum_calib}. The rotation and translation RMSE results of four methods are very small, and these results show that their accuracy is comparable to each other. Inspecting the temporal calibration results, Ours (Midpoint) is almost the same as Kalibr and Basalt. While Ours (Euler) exhibits a larger RMSE (around 2.3 ms), which once again demonstrates the necessity of Midpoint integration.

\subsubsection{Advantages on efficiency}

Comparing the optimization time of different methods, on average, Ours (Midpoint) is 693x faster than Kalibr, and 71x faster than Basalt, offering remarkable advancement in efficiency.

\subsubsection{Impact on VIO accuracy}

The impact of different calibration methods on VIO accuracy is presented in TABLE \ref{tab: vio_tumvi}. This table again shows that our calibration method generates comparable VIO accuracy, as Kalibr or Basalt.

\section{Conclusion}

We propose a novel IMU-Camera calibration method by unleashing the power of discrete-time state representation, pushing the efficiency boundary for the IMU-Camera spatial-temporal calibration. Moreover, the weakness of discrete-time state representation in temporal calibration is tackled with a higher-order IMU preintegration. Experiments demonstrate that our method is extremely faster than the SOTA methods using continuous-time state representation, while maintaining comparable calibration accuracy. This method has the potential to empower higher productivity for commercial products that require massive IMU-Camera factory calibration, such as drones, cellphones and AR glasses.

In the future, our method can be further extended to integrate joint intrinsic refinement for IMU and camera, and accelerate the calibration for multi-visual-inertial systems \cite{yang2024multi}. We are also interested in benchmarking continuous-time and discrete-time state representations for other IMU-aided state estimation tasks \cite{cioffi2022continuous, talbot2025continuous}. Higher-order numerical integration \cite{rk4} could be employed to ensure a fairer comparison for discrete-time state representation.









\bibliographystyle{ieeetr}
\bibliography{bib}




\end{document}